\PassOptionsToPackage{dvipsnames}{xcolor}
\documentclass{fairmeta}
\usepackage{amsmath,amssymb,xspace}
\usepackage{colortbl,tabularx,xurl}
\usepackage[shortlabels,inline]{enumitem}

\setcitestyle{numbers,square,sort&compress}
\hypersetup{
 pdftitle={NativeDrive: Native Multimodal Foundations for Autonomous Driving},
 pdfauthor={Zehua Fan, Qi Jiang, Wei Li, Kai Tang, Yue Jiang, Yujian Li, Bailin Li, Feng Gao, Yan Wang, Kun Zhan},
 pdfdisplaydoctitle=true}


\usepackage[labeled,resetlabels]{multibib}
\newcites{S}{Supplementary References}

\title{NativeDrive: Native Multimodal Foundations for Autonomous Driving}
\author[2,1,\dagger]{Zehua~Fan}
\author[1,\dagger]{Qi~Jiang}
\author[1]{Wei~Li}
\author[1,4]{Kai~Tang}
\author[1,\ddagger]{Yue~Jiang}
\author[1]{Yujian~Li}
\author[1]{Bailin~Li}
\author[2]{Feng~Gao}
\author[3,*]{Yan~Wang}
\author[1,*]{Kun~Zhan}
\affiliation[1]{Li Auto}
\affiliation[2]{Shanghai Jiao Tong University}
\affiliation[3]{Institute for AI Industry Research (AIR), Tsinghua University}
\affiliation[4]{Zhejiang University}
\contribution[\dagger]{Equal contribution}
\contribution[\ddagger]{Project leader}
\contribution[*]{Corresponding authors}
\abstract{Autonomous driving requires semantic understanding and fine-grained scene perception for precise action prediction.
Vision-language-action models (VLAs) inherit semantic priors, but low-dimensional action labels provide limited structural supervision. Some studies extend VLAs with visual generation tasks or adopt video-based world-action models (WAMs), but both approaches introduce appearance-reconstruction overhead, which is less relevant to planning.
Native multimodal models offer an alternative to unify understanding and generation within one backbone. On this basis, we propose NativeDrive, a native encoder-free multimodal driving framework adapting the unified backbone for policy learning and introduce a dense supervision task tailored to driving.
The Native Action Interface reuses the understanding expert’s token-prediction interface for the discrete policy and adapts the generation expert for continuous policy. Driving Semantic Editing reuses the generation expert’s editing capability for dense scene-structure supervision during training, while inference runs only the selected planning policy.
NativeDrive achieves competitive planning performance among unified understanding-and-generation models on NAVSIM v1 and v2, with ablations validating its components' effectiveness.
Code will be released.
}
\date{November 24, 2026}
\begin{document}
\maketitle
\begin{figure*}[!ht]
    \centering
    \includegraphics[width=\textwidth]{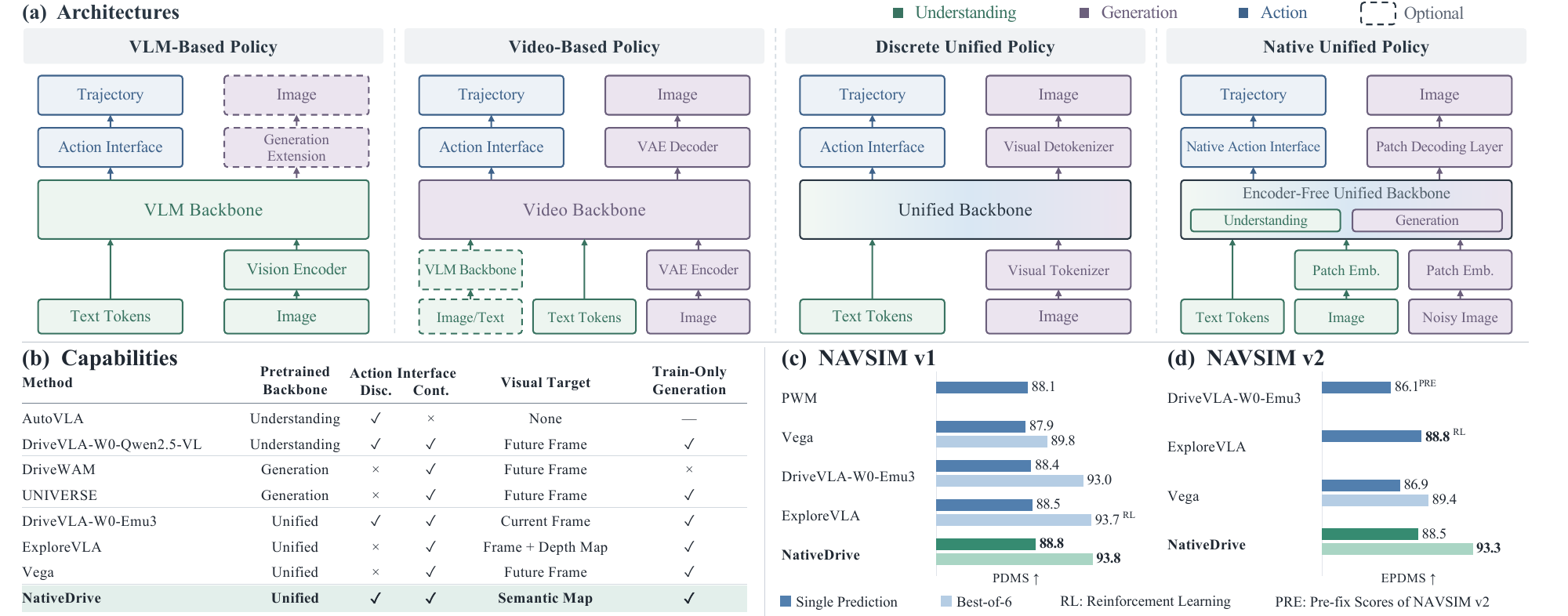}
    \caption{NativeDrive among driving foundation models.
    (a) Four backbone architectures and their action and visual-generation interfaces.
    (b) Key capabilities of representative methods. NativeDrive combines a unified backbone, discrete and continuous action interfaces, and training-only semantic editing.
    (c,d) Unified-model comparisons on NAVSIM v1 (PDMS) and v2 (EPDMS).}
    \label{fig:overview}
\end{figure*}

\FloatBarrier
\section{Introduction}
\label{sec:intro}

Autonomous driving requires models to integrate high-level semantic understanding with fine-grained scene perception for reliable driving actions~\cite{drivevlm,vladbench,drivelm}.
Large pretrained models offer transferable priors for these tasks: vision-language models (VLMs) supply world knowledge and semantic understanding~\cite{autovla,recogdrive}, while video generation models provide motion and temporal dynamics priors for modeling environmental change~\cite{drivewam}.
Transferring these capabilities to driving policies requires a suitable action interface. Low-dimensional trajectory labels alone only weakly supervise learning scene structure~\cite{w0}.
The key question is how to design action interfaces that match the pretrained backbone's modeling mechanisms and dense visual objectives that reflect driving requirements.

Existing approaches primarily build driving policies on vision-language or video generation backbones.
The VLM-based route, shown in the leftmost column of Fig.~\ref{fig:overview}(a), transfers semantic knowledge and instruction understanding into planning through an action interface~\cite{autovla,recogdrive,fsdrive,w0}.
Some methods introduce dense visual supervision by adding an external generation module~\cite{w0} or extending the vocabulary with visual tokens~\cite{w0,fsdrive}.
The video-based route, shown in the second column of Fig.~\ref{fig:overview}(a), exploits pretrained motion and temporal dynamics priors by learning from future-scene generation and action prediction~\cite{drivelaw,drivewam,driveva,universe}.
Prior work uses attention masking to decouple action prediction from future visual tokens, enabling trajectory-only inference~\cite{universe}.
Both routes achieve competitive planning performance.
However, understanding-oriented VLMs typically rely on an extended generation module and lack jointly pretrained understanding and generation foundations~\cite{w0,fsdrive}. Video generation backbones model appearance and dynamics~\cite{drivewam}, but future-frame reconstruction also captures visual details less relevant to planning~\cite{imagewam}.

Unified understanding-and-generation backbones provide another route to transfer both types of priors~\cite{doe1,w0,explorevla,vega}.
In the 3rd column of Fig.~\ref{fig:overview}(a),
some approaches use unified backbones that natively model discrete visual and text tokens to predict future visual tokens~\cite{doe1,w0,explorevla}, either reusing token prediction for discrete actions~\cite{doe1,w0} or extending a continuous trajectory prediction head~\cite{explorevla}.
Recently, native encoder-free architectures enable multimodal understanding and generation via direct image-patch embeddings and pixel prediction~\cite{u1}, reducing reliance on separate visual encoders and intermediate representation alignment.
This design supports jointly developing action interfaces and visual supervision for driving.
Building on this, we propose \textbf{NativeDrive}, jointly designing a policy interface and an auxiliary visual task around the backbone’s native modeling mechanisms.
In the rightmost column of Fig.~\ref{fig:overview}(a), we build discrete and continuous driving policies using the native mechanisms of the understanding and generation experts, respectively. We also use backbone generation for a driving-specific auxiliary visual task.
NativeDrive combines these key attributes: a unified backbone, flexible action interfaces, and training-only visual task as illustrated in Fig.~\ref{fig:overview}(b).

Our contributions are:
\begin{itemize}
\item We propose NativeDrive, which adapts a native unified multimodal backbone to driving policy learning by pairing discrete action prediction with either continuous trajectory generation or training-only semantic editing.
\item We design a Native Action Interface aligning two policy branches with the backbone's native modeling mechanisms. The discrete branch uses the understanding expert to predict quantized action tokens, while the continuous branch adapts the generation expert into a lightweight action head.
\item We introduce Driving Semantic Editing, adapting the generation expert's pretrained editing capability to produce semantic maps. This training-only task provides dense supervision on road layout and traffic participants without additional inference cost.

\item We investigate key design choices for adapting a unified multimodal model to autonomous driving, including action interface, visual task, and joint training. Comparisons and ablations demonstrate the feasibility and effectiveness of the proposed components. NativeDrive achieves 88.8 PDMS on NAVSIM v1~\cite{navsim} and 88.5 EPDMS on NAVSIM v2. Under best-of-6 oracle evaluation, it achieves 93.8 PDMS and 93.3 EPDMS.
\end{itemize}

\section{Related Work}
\label{sec:related}

\textbf{Unified multimodal models} support understanding and content generation within one framework, enabling capabilities to transfer across modalities~\cite{chameleon,emu3,showo,bagel}.
Unified multimodal models can be grouped by whether they use discrete or continuous representations for visual generation.
Chameleon~\cite{chameleon} and Emu3~\cite{emu3} autoregressively model both text and discrete image tokens, whereas Show-o~\cite{showo} combines autoregressive language modeling with discrete diffusion for image generation.
Continuous approaches generate images or videos in latent or pixel space. 
BAGEL~\cite{bagel} combines autoregressive language modeling with continuous image generation through modality-specific experts that share self-attention. Show-o2~\cite{showo2} uses flow matching for both image and video generation.
SenseNova-U1~\cite{u1} unifies visual understanding and generation in an encoder-free architecture with convolutional patch embeddings for input and direct pixel prediction for output.

\begin{figure*}[!t]
  \centering
  \includegraphics[width=\textwidth]{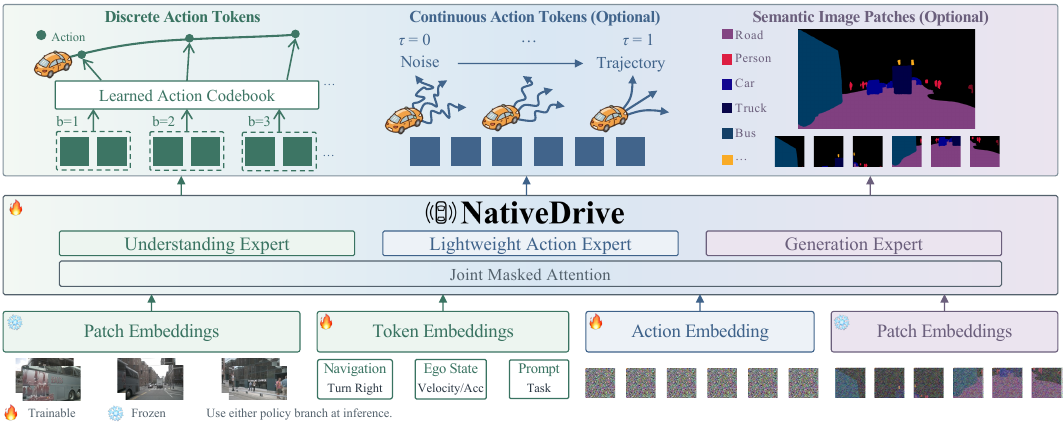}
  \caption{\textbf{NativeDrive framework and policy configurations.} Multi-view images and textual driving context form the shared input for three experts. The understanding expert predicts action tokens that index a learned action codebook for trajectory decoding. The lightweight action expert generates a continuous trajectory from noise, and the generation expert produces a semantic map through training-only Driving Semantic Editing.}
  \label{fig:framework}
\end{figure*}

\textbf{Driving action interfaces} typically use discrete token prediction or continuous trajectory generation. Discrete interfaces tokenize trajectories through per-dimension scalar quantization~\cite{drivinggpt}, motion-primitive clustering~\cite{autovla}, or frequency-domain coding~\cite{w0}.
Doe-1~\cite{doe1} builds on Lumina-mGPT~\cite{lumina_mgpt} and discretizes displacement and yaw, while DrivingGPT~\cite{drivinggpt} uses component-wise binning with a Llama-style autoregressive backbone. AutoVLA~\cite{autovla} constructs physical action tokens through K-disk clustering. DriveVLA-W0~\cite{w0} uses FAST~\cite{fast} to tokenize trajectories in its Emu3~\cite{emu3} and Qwen2.5-VL~\cite{qwen25vl} variants. FAST applies a discrete cosine transform followed by quantization and byte-pair encoding.
Continuous action interfaces use regression, diffusion, or flow matching. DriveVLA-W0~\cite{w0} also connects a lightweight action expert to the backbone through joint attention for query-based regression or flow matching. Vega~\cite{vega} uses a compact action transformer to generate trajectories through diffusion. DriveWAM~\cite{drivewam} and UNIVERSE~\cite{universe} jointly model continuous trajectories and video latents within a generative backbone.
Beyond driving, action representations in embodied policies~\cite{fast,actioncodec} can inform the design of action interfaces for driving policies.

\textbf{Dense visual supervision for driving policies} varies in task design, how visual generation is integrated, and whether action inference requires it. For task design, DriveVLA-W0~\cite{w0} and Vega~\cite{vega} predict future images, FSDrive~\cite{fsdrive} generates future frames annotated with lane dividers and 3D object boxes, and ExploreVLA~\cite{explorevla} jointly predicts future images and depth maps. To incorporate visual generation, DriveWAM~\cite{drivewam} builds on a pretrained video generation backbone, while ImageWAM~\cite{imagewam} reuses pretrained image editing for robot policy learning. DriveVLA-W0~\cite{w0} and FSDrive~\cite{fsdrive} extend understanding-oriented backbones with a generation module and an expanded visual vocabulary, respectively. At inference, FSDrive~\cite{fsdrive}, DriveWAM~\cite{drivewam}, and ImageWAM~\cite{imagewam} rely on visual generation for action prediction. DriveVLA-W0~\cite{w0}, Vega~\cite{vega}, ExploreVLA~\cite{explorevla}, UNIVERSE~\cite{universe}, and Fast-WAM~\cite{fastwam} can predict actions without visual generation.


\section{Method}
\label{sec:method}
In this section, we explore native action adaptation and semantic
supervision for driving with a unified multimodal backbone
(Fig.~\ref{fig:framework}). The \emph{Native Action Interface}
reuses the understanding expert for discrete action token prediction
and adapts the generation expert into a lightweight continuous
action head. The two training configurations depend on how the
generation expert is reused. One jointly trains the discrete and
continuous branches to study how discrete supervision supports
continuous planning, while the other pairs the discrete branch
with training-only \emph{Driving Semantic Editing} for dense
scene-structure supervision.

\subsection{Framework}
\label{sec:framework}
\paragraph{Native Multimodal Backbone.}
We adopt SenseNova-U1~\cite{u1} as our native multimodal backbone. Its encoder-free architecture uses lightweight convolutional patch embeddings for image input and direct pixel prediction for visual output. With multimodal understanding competitive with leading understanding-only models and strong visual generation capabilities, it provides a foundation for transferring both capabilities to driving.

\paragraph{Model Inputs and Outputs.}
At time $t$, the input is $C_t=(\mathcal{I}_t,u_t,s_t)$, where $\mathcal{I}_t$ comprises multi-view visual observations from current and preceding frames. 
The navigation instruction $u_t$ and ego state $s_t$, including velocity and acceleration, are text-serialized and embedded with the system prompt. The primary output is a future ego trajectory:
\begin{equation}
Y_t=\{(x_k,y_k,\psi_k)\}_{k=1}^{H},
\end{equation}
where $(x_k,y_k)$ and $\psi_k$ denote the position and heading relative to the current ego pose, and $H$ counts future waypoints.

\paragraph{Policy Configurations.}
NativeDrive supports two policies based on how the generation expert is reused. The understanding expert provides discrete action prediction in each. The first uses the generation expert for auxiliary semantic editing, while the second adapts it into a lightweight continuous action head. Section~\ref{sec:composition} describes training and inference for each configuration.

\begin{figure}[!t]
  \centering
  \includegraphics[width=0.80\linewidth]{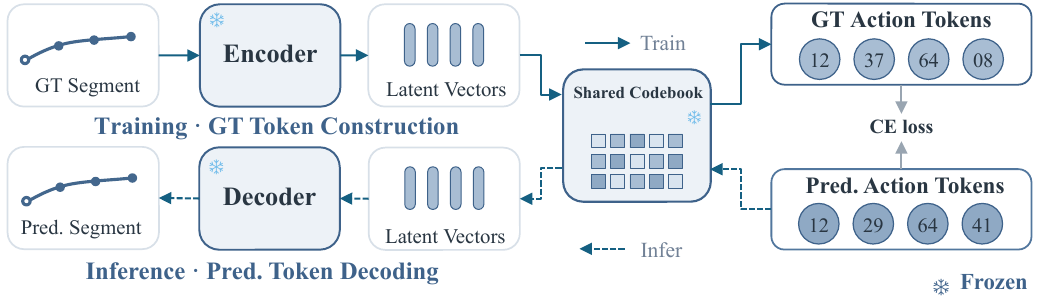}
  \caption{\textbf{Discrete Action Tokenization.} Ground-truth trajectories are encoded and quantized into action tokens, while codebook lookup and decoding convert predicted action tokens into trajectories.}
  \label{fig:actioncodec}
\end{figure}

\subsection{Native Action Interface}
\label{sec:action}
\subsubsection{Discrete Policy Adaptation}
\label{sec:discrete}

\paragraph{Action Tokenization.}
Following the local motion representation in driving action tokenization~\cite{autovla}, we partition each trajectory into non-overlapping segments and express each in its starting ego coordinate frame. Consecutive displacements and wrapped heading increments form:
\begin{equation}
\Delta Y^{(b)}
=
\{(\Delta x_i,\Delta y_i,\Delta\psi_i)\}_{i=1}^{L},
\label{eq:motion-increments}
\end{equation}
where $L$ counts motion samples per segment. We adopt a Perceiver-based encoder--decoder with vector quantization (VQ), originally for robot action sequences~\cite{actioncodec}, and train it to tokenize driving trajectories through reconstruction. Let $X^{(b)}=\operatorname{Norm}(\Delta Y^{(b)})$. The encoder maps each segment to $M$ latent vectors:
\begin{equation}
Z^{(b)}=E_\phi(X^{(b)}).
\label{eq:traj-encode}
\end{equation}
Each latent vector undergoes quantization by nearest-neighbor codebook lookup $\mathcal{E}=\{e_j\}_{j=1}^{K}$:
\begin{equation}
q_i^{(b)}
=
\arg\min_{j\in\{1,\ldots,K\}}
\left\|z_i^{(b)}-e_j\right\|_2^2.
\label{eq:traj-quantize}
\end{equation}
Jointly decoding the selected code vectors reconstructs the normalized segment:
\begin{equation}
\widehat{X}^{(b)}
=
D_\theta(e_{q_1^{(b)}},\ldots,e_{q_M^{(b)}}).
\label{eq:traj-decode}
\end{equation}
We use $M=4$ tokens per $1$\,s segment, latent dimension $d=512$, and codebook size $K=2048$. 
Tokenizer training uses reconstruction and commitment losses~\cite{actioncodec}:
\begin{equation}
\mathcal{L}_{\mathrm{tok}}
=
\left\|X^{(b)}-\widehat{X}^{(b)}\right\|_2^2
+
\frac{\beta}{M}\sum_{i=1}^{M}
\left\|z_i^{(b)}-\operatorname{sg}\!\left(e_{q_i^{(b)}}\right)\right\|_2^2,
\label{eq:tokenizer-loss}
\end{equation}
where $\operatorname{sg}(\cdot)$ denotes the stop-gradient operator, and we set the commitment-loss weight to $\beta=0.25$. Figure~\ref{fig:actioncodec} illustrates how the pretrained tokenizer is used during policy learning and inference; its encoder, codebook, and decoder remain frozen in both stages.

\paragraph{Blockwise Autoregressive Prediction.}
We reuse the understanding expert's native token prediction interface for discrete action prediction. Fine-grained tokenization yields long action sequences; we use blockwise autoregressive (BAR) prediction~\cite{faster} to shorten decoding and reduce error accumulation. We predict $B$ temporal blocks per trajectory sequentially and
the $M$ tokens within each block in parallel. During training, each block is conditioned on the observation context $C$ and preceding ground-truth blocks $Q^{(<b)}$. We minimize the cross-entropy averaged over all $B M$ tokens:
\begin{equation}
\mathcal{L}_{\mathrm{disc}}
= -\frac{1}{BM}
\sum_{b=1}^{B}\sum_{i=1}^{M}
\log p_\theta\!\left(q_i^{(b)} \mid C,Q^{(<b)}\right).
\label{eq:disc}
\end{equation}

\subsubsection{Continuous Policy Adaptation}
\label{sec:continuous}

\paragraph{Lightweight Action Expert.}
We initialize a lightweight action expert from the generation expert to transfer pretrained capabilities for efficient action planning. Following the reduced-width action-expert design~\cite{fastwam}, we retain Transformer depth and attention-head layout while reducing the hidden and feed-forward widths, yielding a compact action expert. Appendix~\ref{app:continuous_implementation} details the layer-wise weight interpolation and expert dimensions for the FM-only configuration. We initialize weights by interpolating the generation weights to match the reduced dimensions, while randomly initializing the action input projection and output head.

\paragraph{Continuous Action Generation.}
We train the action expert with conditional flow matching~\cite{lipman2023flow} using the action-space $v$-loss:
\begin{equation}
\mathcal{L}_{\mathrm{cont}}
= \mathbb{E}\!\left[\operatorname{MSE}
\!\left(v_\theta^{A}, v_\star^{A}\right)\right],
\end{equation}
where $v_\theta^{A}$ and $v_\star^{A}$ denote the predicted and target flow velocities for normalized motion increments. 


\begin{figure}[!h]
  \centering
  \includegraphics[width=0.80\linewidth]{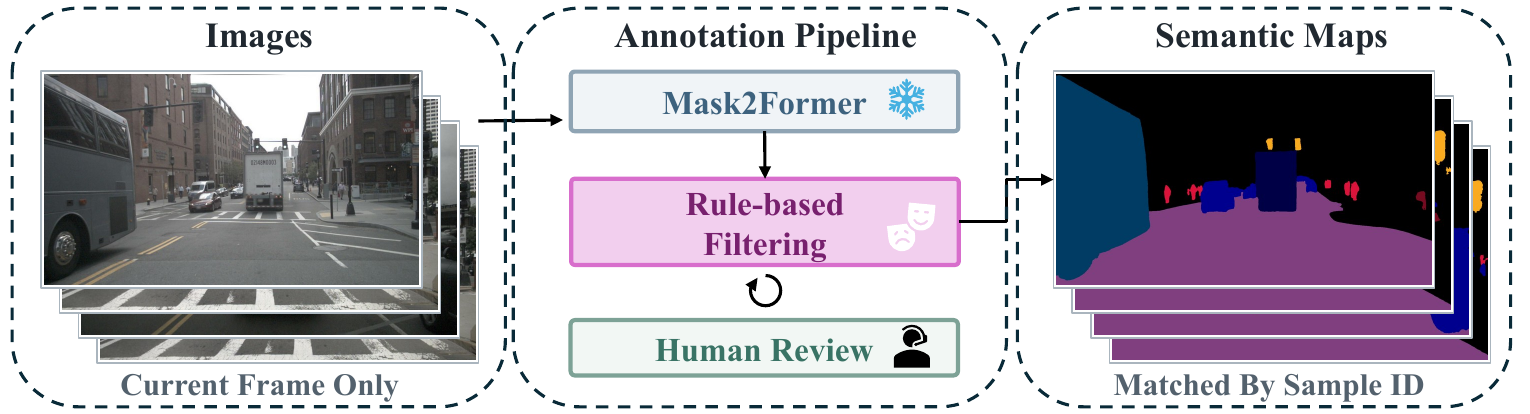}
  \caption{\textbf{Semantic editing data construction.} Frozen Mask2Former~\cite{mask2former}{} predictions are filtered with human review of both the filtering rules and resulting semantic maps.}
  \label{fig:segmentation}
\end{figure}

\subsection{Driving Semantic Editing}
\label{sec:semantic}
\paragraph{Task Design.}
We adapt SenseNova-U1's instruction-guided editing priors~\cite{u1} for semantic segmentation of driving scenes. This task preserves road layout and traffic-agent geometry without predicting future dynamics or textures, while providing structured supervision from pretrained visual editing priors~\cite{imagewam}. Following SenseNova-U1~\cite{u1}, we use $x$-prediction for semantic editing, with the clean semantic map as the target and supervision given by the pixel-space $v$-loss
\begin{equation}
\mathcal{L}_{\mathrm{sem}}
= \mathbb{E}\!\left[\operatorname{MSE}
\!\left(v_\theta^{S}, v_\star^{S}\right)\right],
\end{equation}
where $v_\theta^{S}$ and $v_\star^{S}$ denote the predicted and target flow velocities for the semantic map.

\paragraph{Data Construction.}
Figure~\ref{fig:segmentation} illustrates our offline annotation pipeline for front-camera images across the NAVSIM and Waymo datasets. A frozen Mask2Former~\cite{mask2former} pretrained on Cityscapes~\cite{cityscapes} generates pseudo-labels for eight driving-relevant semantic categories. After removing small components and low-confidence pixels using
rule-based filtering, rejected regions are rendered black. Retained labels are rendered with a fixed RGB palette to form spatially aligned image-semantic map pairs. Human review covers both the filtering rules and the resulting semantic maps, guiding iterative refinement of the rules and reprocessing of failed samples. To preserve original resolutions, inputs are padded to multiples of 32 before prediction and cropped afterward. They are accompanied by a fixed text prompt for instruction-guided editing, while samples without semantic targets omit the editing loss but retain applicable action losses.

\subsection{Training and Inference}
\label{sec:composition}
\begin{figure}[!t]
  \centering
  \includegraphics[width=0.80\linewidth]{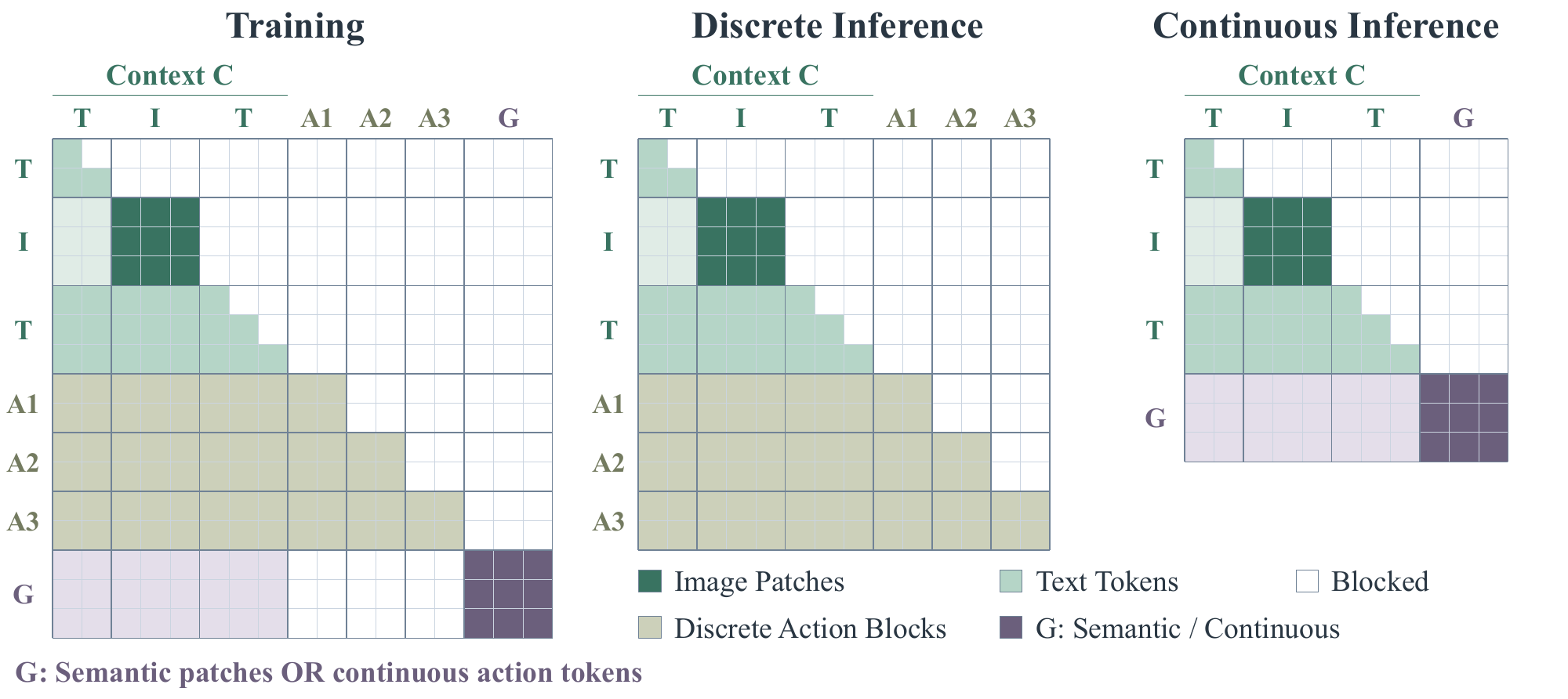}
  \caption{\textbf{Shared training mask and policy-specific inference masks.} Both configurations share a training mask in which discrete action tokens $A$ and generation tokens $G$ attend to context $C$ but cannot attend to each other. $G$ represents either semantic image tokens or continuous action tokens. Discrete inference retains only $C$ and $A$, while continuous inference retains only $C$ and $G$.}
  \label{fig:attention}
\end{figure}

Both configurations use the attention mask in Fig.~\ref{fig:attention}. Discrete action tokens $A$ and generation tokens $G$ attend to the observation context $C$ but cannot attend to each other, and $C$ cannot attend to either target.

\paragraph{Discrete Policy with Semantic Editing.}
We refer to this configuration as \textbf{NativeDrive-D} throughout the remainder of the paper. The understanding expert predicts discrete action tokens, while the generation expert uses semantic image tokens $G$ for Driving Semantic Editing (Sec.~\ref{sec:semantic}). We jointly optimize the discrete policy and semantic editing losses,
\begin{equation}
\mathcal{L}_{\mathrm{total}}
= \mathcal{L}_{\mathrm{disc}}
+ \lambda_{\mathrm{sem}}\mathcal{L}_{\mathrm{sem}},
\end{equation}
where $\lambda_{\mathrm{sem}}=1$ in practice. At inference, only the discrete policy branch runs. Given $C$, it predicts action tokens $A$ blockwise, retrieves their vectors from the learned action codebook, and decodes them into the future trajectory.

\paragraph{Joint Discrete and Continuous Policies.}
We denote this configuration by \textbf{NativeDrive-C}, with $G$ representing continuous action tokens. We initialize the shared vision and understanding modules and the token prediction head from a trained discrete policy, while the continuous action expert retains its initialization from the pretrained generation expert. We jointly optimize the discrete and continuous policy losses,
\begin{equation}
\mathcal{L}_{\mathrm{action}}
= \mathcal{L}_{\mathrm{disc}}
+ \lambda_c\mathcal{L}_{\mathrm{cont}},
\label{eq:action}
\end{equation}
where $\lambda_c=1$ in our experiments. Both losses update the shared observation representation without Knowledge Insulation (KI)~\cite{ki}. Table~\ref{tab:joint_policy} evaluates the effects of KI and discrete policy initialization. At inference, only the continuous policy branch runs. Given the observation context $C$, the action expert denoises the continuous action tokens $G$ from Gaussian noise to generate the future trajectory.

\FloatBarrier

\providecommand{\expTBD}{{\color{red}\textbf{TBD}}}

\newcommand{\AblationFigures}{
\begin{figure}[!htbp]
\centering
\begin{minipage}[t]{0.48\linewidth}
\vspace{0pt}
\centering
\includegraphics[width=\linewidth]{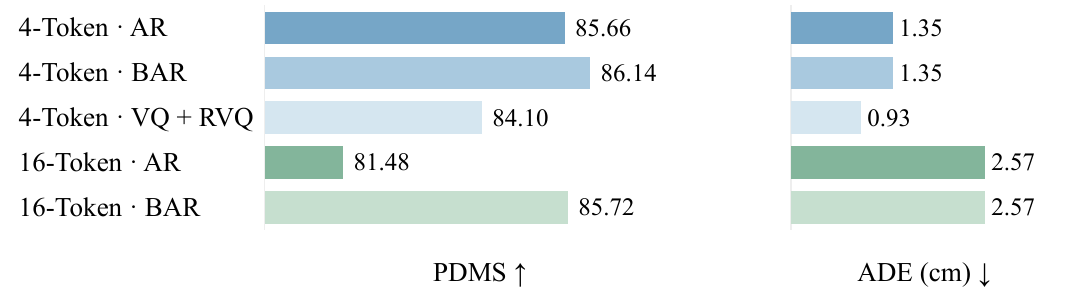}
\caption{\textbf{Token budget and decoding strategy.} All experiments use 360p observations. PDMS measures planning performance, and ADE measures mean trajectory reconstruction error.}
\label{fig:token_budget}
\end{minipage}\hfill
\begin{minipage}[t]{0.48\linewidth}
\vspace{0pt}
\centering
\includegraphics[width=\linewidth]{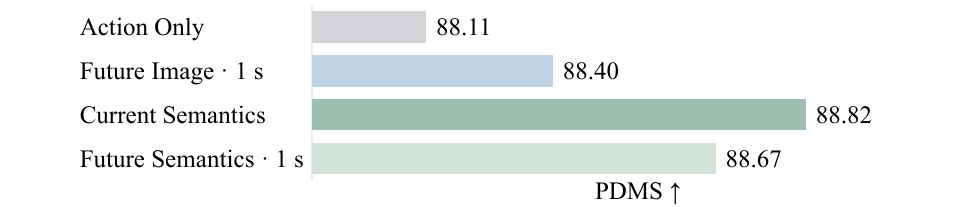}
\caption{\textbf{Auxiliary visual tasks for the discrete policy.} We compare NAVSIM v1 PDMS across supervision settings. Both future prediction tasks use a $1$\,s horizon.}
\label{fig:visual_supervision}
\end{minipage}
\end{figure}
}

\newcommand{\ActionAblationTables}{
\begin{table}[!htbp]
\centering
\begin{minipage}[t]{0.44\linewidth}
\vspace{0pt}
\centering\footnotesize
\caption{\textbf{Discrete action representations on the NAVSIM test set.} NC denotes the no-at-fault collision score, and ADE measures trajectory reconstruction error.}
\label{tab:action_representation}
\setlength{\tabcolsep}{4pt}
\begin{tabular}{lccc}
\toprule
Representation & NC $\uparrow$ & ADE (m) $\downarrow$ & PDMS $\uparrow$ \\
\midrule
Text XY & 96.92 & 0.037 & 82.56 \\
FAST & 97.20 & 0.208 & 83.89 \\
Cluster & 97.91 & 0.049 & 86.59 \\
\rowcolor{gray!12}
\textbf{Learned VQ} {\color{gray}\footnotesize Ours} & \textbf{98.39} & \textbf{0.014} & \textbf{88.11} \\
\bottomrule
\end{tabular}

\end{minipage}\hfill
\begin{minipage}[t]{0.53\linewidth}
\vspace{0pt}
\centering\footnotesize
\setlength{\tabcolsep}{4pt}
\caption{\textbf{Continuous and joint policy training.}
Exp.~1 is the continuous-only baseline, and Exp.~4 corresponds to NativeDrive-C.
Knowledge Insulation (KI)~\cite{ki} blocks continuous-policy gradients to the understanding expert.
Disc. Sup. and Disc. Init. denote discrete policy supervision and initialization from a trained discrete policy, respectively.}
\label{tab:joint_policy}
\begin{tabular}{ccccc}
\toprule
Exp. & KI & Disc. Sup. & Disc. Init. & Continuous PDMS $\uparrow$ \\
\midrule
1 & $\times$     & $\times$     & $\times$     & 84.44 \\
2 & $\checkmark$ & $\checkmark$ & $\times$     & 86.90\,\textcolor{red}{(+2.46)} \\
3 & $\times$     & $\checkmark$ & $\times$     & 88.00\,\textcolor{red}{(+3.56)} \\
\rowcolor{gray!12}
4 & $\times$     & $\checkmark$ & $\checkmark$ & 88.40\,\textcolor{red}{(+3.96)} \\
\bottomrule
\end{tabular}

\end{minipage}
\end{table}
}

\begin{table*}[!t]
\centering
\caption{\textbf{Planning performance on NAVSIM v1.} Methods are grouped by pretrained backbone family. NativeDrive-D achieves the highest PDMS among the listed SFT unified models without driving-domain backbone pretraining.}
\label{tab:navsim_v1}
\begingroup
\small
\renewcommand{\arraystretch}{0.9}
\newcommand{\venue}[1]{{\color{gray}\footnotesize #1}}
\setlength{\tabcolsep}{3pt}
\resizebox{\linewidth}{!}{%
\begin{tabular}{lc*{7}{c}}
\toprule
Method & {\scriptsize\shortstack{Driving\\Pretraining}} & Epochs & NC $\uparrow$ & DAC $\uparrow$ & TTC $\uparrow$ & Comf. $\uparrow$ & EP $\uparrow$ & PDMS $\uparrow$ \\
\midrule
\multicolumn{9}{@{}l}{\textit{VLM-Based Policy}} \\
AutoVLA SFT~\cite{autovla} \venue{NeurIPS 2025} & $\times$ & 5 & 96.9 & 92.4 & 88.1 & 99.9 & 75.8 & 80.5 \\
AutoVLA$^{6,\mathrm{RL}}$~\cite{autovla} \venue{NeurIPS 2025} & $\times$ & 5 & 99.1 & 97.1 & 97.1 & 100.0 & 87.6 & 92.1 \\
FSDrive~\cite{fsdrive} \venue{NeurIPS 2025} & $\checkmark$ & -- & 98.2 & 93.8 & 93.3 & 99.9 & 80.1 & 85.1 \\
ReCogDrive IL~\cite{recogdrive} \venue{ICLR 2026} & $\checkmark$ & 200 & 98.1 & 94.7 & 94.2 & 100.0 & 80.9 & 86.5 \\
DriveVLA-W0 (Qwen)~\cite{w0} \venue{ICLR 2026} & $\checkmark$ & $\approx 1.9$ & 98.3 & 93.5 & 94.8 & 100.0 & 79.1 & 85.3 \\
\midrule
\multicolumn{9}{@{}l}{\textit{Video-Based Policy}} \\
DriveLaW~\cite{drivelaw} \venue{arXiv 2025} & $\checkmark$ & $\approx 82.0$ & 99.0 & 97.1 & 96.7 & 100.0 & 81.3 & 89.1 \\
DriveDreamer-Policy~\cite{drivedreamerpolicy} \venue{arXiv 2026} & $\times$ & $\approx 31.1$ & 98.4 & 97.1 & 95.1 & 100.0 & 83.5 & 89.2 \\
DriveWAM~\cite{drivewam} \venue{NeurIPS 2026} & $\times$ & $\approx 46.6$ & 98.3 & 98.1 & 95.2 & 100.0 & 84.3 & 90.1 \\
DriveVA~\cite{driveva} \venue{ECCV 2026} & $\times$ & $\approx 77.7$ & 99.0 & 97.3 & 98.4 & 100.0 & 83.2 & 90.5 \\
UNIVERSE~\cite{universe} \venue{NeurIPS 2026} & $\times$ & $\approx 77.7$ & 99.1 & 97.6 & 98.5 & 100.0 & 83.6 & 91.0 \\
\midrule
\multicolumn{9}{@{}l}{\textit{Unified Model}} \\
Vega~\cite{vega} \venue{arXiv 2026} & $\times$ & $\approx 15.5$ & 98.9 & 95.3 & 96.1 & 100.0 & 81.6 & 87.9 \\
Vega$^{6}$~\cite{vega} \venue{arXiv 2026} & $\times$ & $\approx 15.5$ & 99.2 & 96.6 & 96.9 & 100.0 & 83.4 & 89.8 \\
PWM~\cite{pwm} \venue{NeurIPS 2025} & $\checkmark$ & 20 & 98.6 & 95.9 & 95.4 & 100.0 & 81.8 & 88.1 \\
DriveVLA-W0(Emu3)~\cite{w0} \venue{ICLR 2026} & $\checkmark$ & $\approx 1.9$ & 98.7 & 96.2 & 95.5 & 100.0 & 82.2 & 88.4 \\
DriveVLA-W0(Emu3)$^{6}$~\cite{w0} \venue{ICLR 2026} & $\checkmark$ & $\approx 1.9$ & 99.3 & 97.4 & 97.0 & 99.9 & 88.3 & 93.0 \\
ExploreVLA SFT~\cite{explorevla} \venue{ECCV 2026} & $\checkmark$ & 25 & 98.7 & 96.2 & 95.7 & 100.0 & 82.3 & 88.5 \\
ExploreVLA$^{6,\mathrm{RL}}$~\cite{explorevla} \venue{ECCV 2026} & $\checkmark$ & 30 & 99.4 & 98.9 & 98.3 & 99.7 & 88.3 & 93.7 \\
\rowcolor{gray!12}
\textbf{NativeDrive-C} \venue{Ours} & $\times$ & 10 & 98.4 & 96.1 & 95.0 & 100.0 & 82.9 & 88.4 \\
\rowcolor{gray!12}
\textbf{NativeDrive-C}$^{6}$ \venue{Ours} & $\times$ & 10 & 98.8 & 97.8 & 96.4 & 100.0 & 85.6 & 91.1 \\
\rowcolor{gray!12}
\textbf{NativeDrive-D} \venue{Ours} & $\times$ & 5 & 98.5 & 96.6 & 95.2 & 100.0 & 83.2 & 88.8 \\
\rowcolor{gray!12}
\textbf{NativeDrive-D}$^{6}$ \venue{Ours} & $\times$ & 5 & 99.2 & 98.7 & 97.7 & 99.8 & 89.2 & \textbf{93.8} \\
\bottomrule
\end{tabular}
}
\endgroup
\par\vspace{2pt}
\noindent\begin{minipage}{\linewidth}\footnotesize
\textit{Driving Pretraining}: driving-domain pre-training before SFT;
$\approx$: step-based NAVSIM-equivalent epochs. $\mathrm{RL}$ indicates reinforcement learning after supervised fine-tuning. $^{6}$ denotes best-of-6 evaluation.
NativeDrive-D samples six trajectories per scene with temperature
$1.1$, top-$p=0.95$, and no top-$k$ truncation;
NativeDrive-C uses six distinct Gaussian initializations and 10 Euler steps.
For both, the oracle maximizes per-scene PDMS before averaging,
with submetrics from the same selected trajectory.
\end{minipage}
\end{table*}

\section{Experiments}
\label{sec:experiments}
In this section, we compare NativeDrive with representative state-of-the-art planners, evaluate its key components through ablations, and present visualization results.

\subsection{Experimental Settings}
\paragraph{Datasets and Benchmarks.}
We evaluate NativeDrive on NAVSIM~\cite{navsim} using its official splits and protocols. We report PDMS on v1 and single-stage EPDMS on v2 \texttt{navtest}~\cite{navsimmetrics}. Each input includes current and preceding frames from the front, front-left, and front-right cameras, sampled $0.5$\,s apart. The current front-view image is resized to 720p, and all remaining images to 360p. We predict ten future waypoints over a $5$\,s horizon at 2\,Hz. All experiments use these settings, unless otherwise specified.

\paragraph{Tokenizer and Policy Training.}
We train the VQ action tokenizer on trajectories sampled with equal probability from NAVSIM~\cite{navsim} and Waymo E2E~\cite{wode2e}. For tokenizer training, we use AdamW~\cite{adamw} for 50,000 steps on 8 GPUs with a learning rate of $2\times10^{-4}$ and a global batch size of 16,384. For policy training, we fine-tune SenseNova-U1-8B-MoT-SFT~\cite{u1} with AdamW, freezing the observation image embedding module and action tokenizer. We use a global batch size of 32, weight decay of $0.1$, and a 50-step warm-up to a peak learning rate of $5\times10^{-5}$, followed by cosine decay to $10^{-6}$. We train the discrete policy for 5 epochs on 8 H20 GPUs, with or without semantic editing. For joint training, we initialize the shared modules and token prediction head from the discrete-only checkpoint and the continuous action head from the pretrained generation expert, then train both branches for 10 epochs.

\subsection{Main Results}
\paragraph{NAVSIM v1.}
NativeDrive-C and NativeDrive-D score $88.4$ and $88.8$ PDMS on NAVSIM v1, respectively, rising to $91.1$ and $93.8$ under best-of-six oracle selection (Table~\ref{tab:navsim_v1}). Without driving-domain backbone pretraining, NativeDrive-D leads the listed SFT unified models after just five policy-training epochs and achieves the highest listed best-of-six score.

\paragraph{NAVSIM v2.}
On NAVSIM v2 \texttt{navtest}, NativeDrive-D delivers strong performance among unified models under the post-fix protocol, achieving $88.5$ EPDMS in single-trajectory evaluation and an outstanding $93.3$ EPDMS under best-of-six oracle selection. NativeDrive-C achieves competitive scores of $88.1$ and $91.0$ EPDMS under the same settings (Table~\ref{tab:navsim_v2}).

\begin{table*}[t]
\centering
\providecommand{\venue}[1]{{\color{gray}\footnotesize #1}}
\caption{\textbf{Planning performance on NAVSIM v2 \texttt{navtest}.} Methods are grouped by pretrained backbone family, with pre-fix and post-fix scores reported separately. After five epochs of policy training without driving-domain backbone pretraining, NativeDrive-D achieves $88.5$ EPDMS and $93.3$ EPDMS under best-of-six oracle selection.}
\label{tab:navsim_v2}
\label{tab:additional_benchmarks}
\begingroup
\small
\setlength{\tabcolsep}{0.5pt}
\renewcommand{\arraystretch}{0.9}
\begin{tabularx}{\textwidth}{l*{11}{>{\centering\arraybackslash}X}}
\toprule
Method & NC & DAC & DDC & TLC & EP & TTC & LK & HC & EC & {\scriptsize\shortstack{EPDMS$^*$\\Pre-fix}} & {\scriptsize\shortstack{EPDMS\\Post-fix}} \\
\midrule
\multicolumn{12}{@{}l}{\textit{VLM-Based Policy}} \\
ReCogDrive~\cite{recogdrive} & 98.3 & 95.2 & 99.5 & 99.8 & 87.1 & 97.5 & 96.6 & 98.3 & 86.5 & 83.6 & -- \\
\midrule
\multicolumn{12}{@{}l}{\textit{Video-Based Policy}} \\
Epona~\cite{epona}$^{a}$ & 97.1 & 95.7 & 99.3 & 99.7 & 88.6 & 96.3 & 97.0 & 98.0 & 67.8 & -- & 85.1 \\
DreamerAD~\cite{dreamerad} & 98.0 & 97.2 & 99.5 & 99.8 & 87.8 & 97.4 & 97.5 & 98.3 & 72.4 & -- & 87.7 \\
DriveLaW~\cite{drivelaw}$^{b}$ & 98.7 & 96.9 & 99.6 & 99.8 & 87.5 & 98.3 & 97.6 & 98.4 & 77.4 & -- & 88.6 \\
DriveDreamer-Policy~\cite{drivedreamerpolicy} & 98.4 & 97.1 & 99.5 & 99.9 & 87.9 & 97.7 & 97.6 & 98.3 & 79.4 & -- & 88.7 \\
\midrule
\multicolumn{12}{@{}l}{\textit{Unified Model}} \\
DriveVLA-W0~\cite{w0} & 98.5 & 99.1 & 98.0 & 99.7 & 86.4 & 98.1 & 93.2 & 97.9 & 58.9 & 86.1 & -- \\
ExploreVLA$^{\mathrm{RL}}$~\cite{explorevla} & 98.8 & 96.2 & 99.6 & 99.8 & 87.1 & 98.2 & 97.8 & 98.3 & 86.8 & -- & 88.8 \\
Vega~\cite{vega} & 98.9 & 95.3 & 99.4 & 99.9 & 87.0 & 98.4 & 96.5 & 98.3 & 76.3 & -- & 86.9 \\
Vega$^{6}$~\cite{vega} & 99.2 & 96.6 & 99.5 & 99.9 & 87.5 & 98.7 & 97.4 & 98.4 & 84.5 & -- & 89.4 \\
\rowcolor{gray!12}
\textbf{NativeDrive-C} \venue{Ours}& 98.4 & 95.9 & 99.6 & 99.8 & 87.2 & 97.7 & 97.3 & 98.3 & 86.5 & -- & 88.1 \\
\rowcolor{gray!12}
\textbf{NativeDrive-C}$^{6}$ \venue{Ours}& 98.8 & 97.8 & 99.7 & 99.9 & 88.3 & 98.2 & 98.1 & 98.4 & 89.7 & -- & 91.0 \\
\rowcolor{gray!12}
\textbf{NativeDrive-D} \venue{Ours}& 98.5 & 96.6 & 99.5 & 99.8 & 87.3 & 97.7 & 97.3 & 98.3 & 85.3 & -- & 88.5 \\
\rowcolor{gray!12}
\textbf{NativeDrive-D}$^{6}$ \venue{Ours}& 99.3 & 99.0 & 99.7 & 99.9 & 89.7 & 98.8 & 98.6 & 98.3 & 90.5 & -- & \textbf{93.3} \\
\bottomrule
\end{tabularx}
\endgroup
\vspace{2pt}
\begin{minipage}{\textwidth}\footnotesize
EPDMS$^*$ and EPDMS denote evaluation before and after the official human-filter fix, respectively. $6$ denotes best-of-six evaluation. NC: no at-fault collision; DAC: drivable-area compliance; DDC: driving-direction compliance; TLC: traffic-light compliance; EP: ego progress; TTC: time-to-collision; LK: lane keeping; HC: history comfort; EC: extended comfort.
$^{a}$: results reported in DreamerAD~\cite{dreamerad}; $^{b}$: results reported in RISE~\cite{rise}. --: unavailable. NativeDrive-D$^6$: six samples per scene, temperature $1.3$, top-$p=0.95$, top-$k=50$; the oracle maximizes corrected EPDMS per scene before averaging, with all submetrics taken from that same candidate. NativeDrive-C$^6$: the same as NAVSIM v1.
\end{minipage}
\end{table*}

\subsection{Ablation Studies}
We conduct ablation studies on the \texttt{navtest} set to evaluate discrete action representations, joint discrete--continuous policy training, and visual supervision. All settings remain fixed except the investigated factor.

\paragraph{Action Tokenization.}
Table~\ref{tab:action_representation} compares four discrete action representations under the same backbone and training settings. Cluster uses motion prototypes~\cite{autovla}, FAST uses frequency-domain tokens~\cite{fast}, and Text XY represents waypoint coordinates as text~\cite{autovlad}. Learned VQ achieves the highest PDMS of $88.11$, outperforming the others. Across the four representations, gains in reconstruction fidelity do not always translate into comparable gains in PDMS.



To further examine the effects of token budget and decoding strategy, we compare AR and BAR at $4$ and $16$ tokens/s and include a VQ+RVQ configuration at $4+4$ tokens/s following ActionCodec~\cite{actioncodec}. As shown in Fig.~\ref{fig:token_budget}, BAR outperforms AR by $0.48$ and $4.24$ PDMS points at the two budgets, respectively. The larger gain at $16$ tokens/s suggests that parallel prediction within each block helps limit error accumulation in longer sequences. With BAR, increasing the budget from $4$ to $16$ tokens/s substantially raises reconstruction error and lowers PDMS by $0.42$ points. Adding RVQ improves reconstruction but lowers PDMS to $84.10$, suggesting that single-level VQ already provides sufficient fidelity and that predicting an extra block may hurt planning. Since reconstruction error alone does not reliably predict planning performance, we adopt single-level VQ with BAR at $4$ tokens/s, which achieves the highest PDMS.
\paragraph{Joint Policy Training.}
We evaluate how discrete supervision, gradient sharing, and policy initialization affect continuous planning (Table~\ref{tab:joint_policy}). Adding discrete supervision improves PDMS by $3.56$ points (Exps.~1 and~3). Removing Knowledge Insulation (KI)~\cite{ki} improves PDMS by $1.10$ points (Exps.~2 and~3), possibly because both policy losses can refine shared observation features in the backbone's unified representation space. Initializing from a trained discrete policy adds $0.40$ points (Exps.~3 and~4). The best configuration, NativeDrive-C (Exp.~4), achieves $88.40$ PDMS, $3.96$ points above the continuous-only baseline.
\ActionAblationTables
\paragraph{Visual Supervision.}
To assess how the choice of auxiliary visual task affects discrete planning, we compare an action-only baseline with three alternatives in Fig.~\ref{fig:visual_supervision}: future-frame prediction, current-frame semantic editing, and future semantic prediction. Both future-prediction tasks use a $1$\,s horizon. All three tasks improve on the action-only baseline of $88.11$ PDMS. Current-frame semantic editing achieves the highest PDMS ($88.82$), a gain of $0.71$ points. At the same prediction horizon, future semantic prediction outperforms future-frame prediction ($88.67$ vs.\ $88.40$), suggesting that supervision focused on scene structure is more useful for planning than detailed appearance reconstruction. Moving semantic supervision from the current frame to a future frame slightly reduces PDMS. This may reflect the limited emphasis on motion modeling during SenseNova-U1 pretraining~\cite{u1} and the absence of driving-specific pretraining for the generation task.
\AblationFigures

\subsection{Qualitative Visualization}

\begin{figure}[!htbp]
\centering
\includegraphics[width=\linewidth]{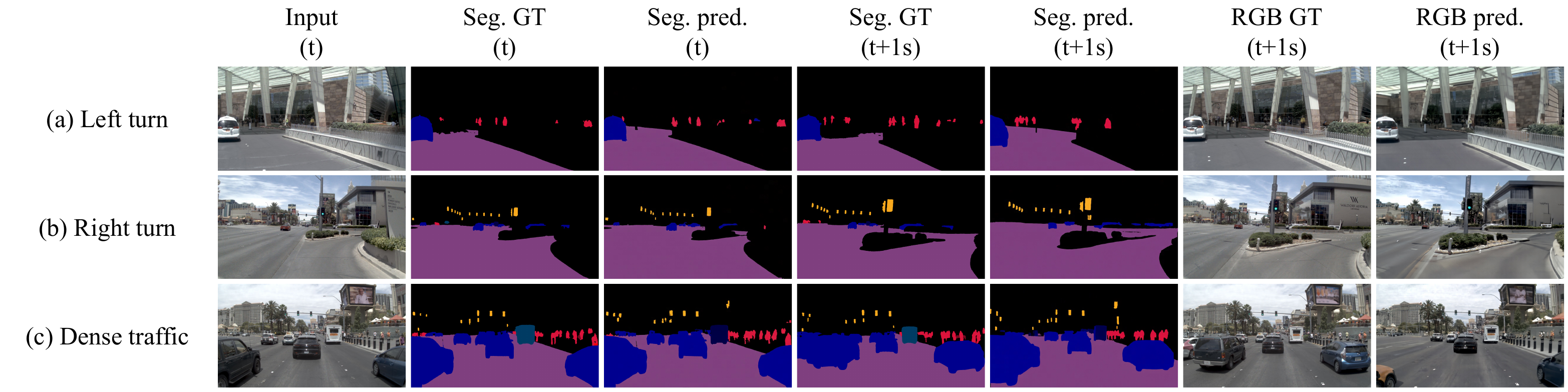}
\caption{Auxiliary visual predictions on shared NAVSIM validation scenes: (a) left turn, (b) right turn, and (c) dense traffic. Rows correspond to the three scenes. From left to right: current front-camera image, current semantic target and prediction, semantic target and prediction at $t+1$\,s, and RGB target and prediction at $t+1$\,s. Each task uses its respective fine-tuned model. Semantic columns labeled ``GT'' contain offline pseudo-labels; RGB targets are recorded camera frames.}
\label{fig:visual_prediction}
\end{figure}

\paragraph{Auxiliary Visual Prediction.}
Figure~\ref{fig:visual_prediction} compares 3 separately fine-tuned auxiliary-task variants' visual outputs on shared NAVSIM validation scenes: current-frame semantic reconstruction, future semantic prediction, and future image prediction. Both future tasks use $1$\,s horizons. We visualize 3 shared scenes covering a left turn, a right turn, and dense traffic. Current-frame semantic predictions recover the broad road layout and prominent traffic participants. Future semantic predictions capture coarse changes in projected positions, while discrepancies remain in object boundaries, small participants, and class assignments. Future image predictions preserve recognizable scene appearance but differ in vehicle placement and scene geometry from the recorded future frames. These examples illustrate the distinct visual targets in Figure~\ref{fig:visual_supervision}; the semantic targets emphasize driving-relevant structure, whereas image prediction additionally requires appearance synthesis. Action and visual generation tokens are mutually masked: visual prediction does not use predicted actions, and action prediction does not use generated visual tokens.

\section{Conclusion}
\label{sec:conclusion}
We presented NativeDrive, a driving framework transferring native multimodal understanding and generation to planning. Its Native Action Interface supports learned discrete action tokens and a continuous policy initialized from the generation expert, while Driving Semantic Editing provides structure-focused visual supervision during training. NAVSIM ablations demonstrate the benefits of learned action representations, blockwise prediction, joint policy training, and current-frame semantic supervision. NativeDrive plans without executing auxiliary visual generation at inference, providing a practical means to reuse unified multimodal priors for driving.
\FloatBarrier
\bibliographystyle{ieeenat_fullname}
\bibliography{main}
\clearpage
\beginappendix
\let\NativeDriveOriginalCiteS\citeS
\renewcommand{\citeS}[1]{%
  \begingroup
  \renewcommand{\citenumfont}[1]{S##1}%
  \NativeDriveOriginalCiteS{#1}%
  \endgroup
}

\setcounter{figure}{0}
\setcounter{table}{0}
\renewcommand{\thefigure}{S\arabic{figure}}
\renewcommand{\thetable}{S\arabic{table}}
\renewcommand{\theHfigure}{supp.\arabic{figure}}
\renewcommand{\theHtable}{supp.\arabic{table}}

\section{Continuous Action Expert Adaptation}
\label{app:continuous_implementation}

We retain the generation expert's Transformer depth and attention configuration while reducing the hidden width from 4096 to 768 and the feed-forward intermediate width from 12288 to 3072. We replace its visual interfaces with an action input projection ($3\rightarrow768$) and a velocity output head ($768\rightarrow3$) for three-dimensional motion increments. Pretrained weights are adapted independently within each corresponding Transformer layer through sequential one-dimensional linear interpolation along the output and input axes, skipping unchanged dimensions, while the action input projection and output head are randomly initialized. The same continuous-expert architecture is used in joint discrete--continuous training; NativeDrive-C additionally uses discrete-policy checkpoint initialization (Table~\ref{tab:joint_policy}).

\section{Supplementary Experiments and Results}
\label{app:additional_experiments}

Our supplementary experiments cover visual supervision integration and generative pretraining, robustness across training seeds, and evaluation on Waymo-E2E.

\subsection{Generation Design and Initialization}
\label{app:native_generation_ablations}

\begin{table}[!htbp]
\centering
\begin{minipage}[t]{0.48\linewidth}
\vspace{0pt}
\centering\footnotesize
\caption{\textbf{Visual supervision integration and generative pretraining.} PDMS on NAVSIM v1 \texttt{navtest}. Panels (a) and (b) compare supervision designs and generation-expert initialization, respectively, with other settings fixed.}
\label{tab:app_generation_design}
\begin{tabular}{lr}
\toprule
Configuration & PDMS $\uparrow$ \\
\midrule
\multicolumn{2}{l}{\textit{(a) Visual supervision integration}} \\
Extended visual vocabulary & 88.28 \\
External diffusion module & 88.28 \\
\rowcolor{gray!12}
\textbf{Native generation expert} & \textbf{88.82} \\
\midrule
\multicolumn{2}{l}{\textit{(b) Generation-expert initialization}} \\
Random initialization & 88.48 \\
\rowcolor{gray!12}
\textbf{Pretrained initialization} & \textbf{88.82} \\
\bottomrule
\end{tabular}

\end{minipage}\hfill
\begin{minipage}[t]{0.48\linewidth}
\vspace{0pt}
\centering\footnotesize
\setlength{\tabcolsep}{2pt}
\caption{\textbf{Sensitivity to training seeds on NAVSIM v1.} Scores are PDMS ($\uparrow$); the last column reports the mean and sample standard deviation across three seeds. Both future targets are $1$\,s ahead.}
\label{tab:app_training_seeds}
\begin{tabular}{lrrrc}
\toprule
Supervision & Seed 41 & Seed 42 & Seed 43 & Mean $\pm$ s.d. \\
\midrule
Action only & 88.21 & 88.11 & 88.37 & $88.23\pm0.13$ \\
\rowcolor{gray!12}
+ Current semantic & 88.44 & 88.82 & 88.65 & $\mathbf{88.64\pm0.19}$ \\
+ Future semantic & 88.39 & 88.67 & 88.45 & $88.50\pm0.15$ \\
+ Future image & 88.47 & 88.40 & 88.62 & $88.50\pm0.11$ \\
\bottomrule
\end{tabular}

\end{minipage}
\end{table}

\paragraph{Visual supervision integration.}
We compare three visual supervision designs. Extended visual vocabulary follows FSDrive~\cite{fsdrive} by adding discrete visual tokens for next-token prediction. External diffusion module follows DriveVLA-W0~\cite{w0} by attaching a latent diffusion module conditioned on the understanding expert's visual features. Native generation expert reuses the backbone's pretrained generation expert and existing visual interface. The first two designs achieve $88.28$ PDMS, while native generation achieves $88.82$, a gain of $0.54$ points (Table~\ref{tab:app_generation_design}(a)), supporting native expert reuse for semantic supervision.

\paragraph{Generation-expert initialization.}
We compare pretrained and random initialization of the native semantic-generation expert while keeping its architecture and training settings fixed. In both cases, the understanding expert retains its pretrained initialization and remains trainable. Pretrained generation weights improve PDMS from $88.48$ to $88.82$ ($+0.34$), supporting the benefit of generative pretraining for downstream planning (Table~\ref{tab:app_generation_design}(b)).

\subsection{Robustness Across Training Seeds}
\label{app:seed_robustness}

We repeat the visual-supervision ablation on NAVSIM v1 with training seeds 41, 42, and 43 to assess whether the gains persist across runs. Table~\ref{tab:app_training_seeds} reports individual scores and their mean and sample standard deviation. All auxiliary visual tasks improve over action-only training across the evaluated seeds. Current semantic editing achieves the highest mean PDMS, exceeding the action-only baseline by $0.41$ points, with positive gains at every seed supporting its consistent benefit to discrete planning.

\begin{table*}[t]
\centering\small
\setlength{\tabcolsep}{3pt}
\newcommand{\waymovenue}[1]{{\color{gray}\scriptsize #1}}
\caption{\textbf{Open-loop planning on Waymo-E2E.} RL denotes reinforcement learning; ``--'' indicates unavailable entries.}
\label{tab:waymo}
\resizebox{\linewidth}{!}{%
\begin{tabular}{lccccccc}
\toprule
& & \multicolumn{3}{c}{Validation} & \multicolumn{3}{c}{Test} \\
\cmidrule(lr){3-5}\cmidrule(l){6-8}
Method / configuration & RL & ADE@3s (m) $\downarrow$ & ADE@5s (m) $\downarrow$ & RFS $\uparrow$ & ADE@3s (m) $\downarrow$ & ADE@5s (m) $\downarrow$ & RFS $\uparrow$ \\
\midrule
Human driver~\citeS{apppoutine} & -- & -- & -- & 8.13 & -- & -- & -- \\
\midrule[0.4pt]
\multicolumn{8}{@{}l}{\textit{Conventional End-to-End Policy}} \\
VAD~\citeS{appvad} \waymovenue{ICCV 2023} & $\times$ & 3.19 & 5.81 & 4.45 & -- & -- & -- \\
UniAD~\citeS{appuniad} \waymovenue{CVPR 2023} & $\times$ & 6.50 & 10.81 & 5.78 & -- & -- & -- \\
UniPlan~\citeS{appuniplan} \waymovenue{Challenge Report 2025} & $\times$ & -- & -- & -- & 1.27 & 2.84 & 7.78 \\
\midrule
\multicolumn{8}{@{}l}{\textit{VLM-Based Policy}} \\
NaiveEMMA~\cite{wode2e} \waymovenue{arXiv 2025} & $\times$ & -- & -- & -- & 1.32 & 3.02 & 7.53 \\
AutoVLA~\cite{autovla} \waymovenue{NeurIPS 2025} & $\checkmark$ & -- & -- & -- & 1.35 & 2.96 & 7.56 \\
NoRD~\citeS{appnord} \waymovenue{CVPR 2026} & $\checkmark$ & -- & -- & -- & 1.25 & 2.89 & 7.71 \\
HMVLM~\citeS{apphmvlm} \waymovenue{arXiv 2025} & $\times$ & -- & -- & -- & 1.33 & 3.07 & 7.74 \\
MindVLA-U1 (SFT)~\citeS{appmindvla} \waymovenue{arXiv 2026} & $\times$ & 0.89 & 2.11 & 7.92 & -- & -- & -- \\
Qwen-Drive-1.0 (SFT)~\citeS{appqwendrive} \waymovenue{arXiv 2026} & $\times$ & 0.99 & 2.31 & 7.95 & 1.19 & 2.65 & 7.78 \\
\midrule
\multicolumn{8}{@{}l}{\textit{Unified Model}} \\
Action only (discrete) & $\times$ & 1.20 & 2.77 & 7.89 & 1.34 & 3.05 & 7.75 \\
+ Semantic editing (NativeDrive-D) & $\times$ & 1.17 & 2.65 & 7.83 & 1.34 & 2.93 & 7.75 \\
\rowcolor{gray!12}
+ Joint FM (NativeDrive-C) & $\times$ & 1.12 & 2.45 & \textbf{7.95} & 1.26 & 2.70 & \textbf{7.86} \\
\bottomrule
\end{tabular}
}
\end{table*}

\subsection{Additional Waymo-E2E Experiments}
\label{app:waymo}

We evaluate NativeDrive's generalization to challenging long-tail scenarios on Waymo-E2E~\cite{wode2e}. We compare its two variants with the action-only baseline and other representative methods.

\paragraph{Dataset and Evaluation.}
We report ADE at $3$ and $5$\,s and Rater Feedback Score (RFS). Validation ADE follows the official protocol with predictions interpolated to $4$\,Hz, while test metrics are obtained from the official evaluation server. Given the limited training data, we jointly train on Waymo-E2E, NAVSIM, and nuScenes, keeping other training settings unchanged.

\paragraph{Comparison with Other Methods.}
NativeDrive-C performs best among our variants on both splits, improving validation/test RFS by $0.06/0.11$ and reducing ADE@5s by $0.32/0.35$\,m over discrete action-only planning. Among the listed methods without RL, it achieves state-of-the-art RFS, with the highest test score of $7.86$ and a tied highest validation score of $7.95$ (Table~\ref{tab:waymo}).

\paragraph{Complementary Strengths of the Two Variants.}
NativeDrive-D provides larger planning gains on NAVSIM v1, while NativeDrive-C shows broader improvements on Waymo-E2E. This may reflect their complementary strengths in scene-structure perception and continuous trajectory refinement, respectively.
{
\small
\setlength{\bibsep}{1pt}
\bibliographystyleS{ieeenat_fullname}
\bibliographyS{appendix}
}

\end{document}